\documentclass{article}

\usepackage{iclr2025_conference,times}

\usepackage{hyperref}
\usepackage{url}

\usepackage{amsmath,amssymb}
\usepackage{booktabs}
\usepackage{graphicx}
\usepackage{microtype}
\usepackage{multirow}
\usepackage{enumitem}
\usepackage{xcolor}
\usepackage{float} 

\usepackage{natbib}
\hypersetup{
    colorlinks=true,
    linkcolor=black,
    citecolor=black,
    urlcolor=blue
}

\newcommand{\marketbench}{\textsc{Market-Bench}}

\title{\marketbench: Evaluating Large Language Models on Introductory Quantitative Trading and Market Dynamics}

\author{
\textbf{Abhay Srivastava}$^{1}$ \quad
\textbf{Sam Jung}$^{1}$\thanks{Correspondence to: \texttt{samj@afterquery.com}.} \quad
\textbf{Spencer Mateega}$^{1}$ \\
$^{1}$AfterQuery
}
\iclrfinalcopy

\begin{document}

\maketitle

\begin{abstract}
We introduce \textbf{\marketbench}, a benchmark that evaluates large language models (LLMs) on \emph{introductory quantitative trading tasks} by asking them to construct executable backtesters from natural-language strategy descriptions and market assumptions. Each instance specifies one of three canonical strategies—scheduled trading on Microsoft (NASDAQ: MSFT), pairs trading on Coca-Cola (NASDAQ: KO) and Pepsi (NASDAQ: PEP), or delta hedging on MSFT—and models must produce code whose P\&L, drawdown, and position paths match a verifiable reference implementation. We assess thirteen state-of-the-art models using a multi-round evaluation that separates structural reliability (whether the backtest runs) from numerical accuracy (mean absolute error of the backtest metrics), assigning failed outputs a duplicated-metrics baseline MAE. While most models reliably execute the simplest strategy (average Executable Passes of $4.08$ out of $5$ rounds), errors vary by orders of magnitude across models and tasks: Gemini 3 Pro and Claude 4.5 Sonnet combine strong reliability with low error on simpler strategies, GPT-5.2 achieves strong overall performance with perfect executability, GPT-5.1 Codex-Max achieves the lowest best-run error on the easiest task, and Qwen3 Max attains perfect executability yet sometimes produces inaccurate P\&L paths. These results show that current LLMs can scaffold basic trading infrastructure but still struggle to reason robustly about prices, inventory, and risk; we release \marketbench{} and a public leaderboard at: \url{https://marketbench.ai}.
\end{abstract}

\section{Introduction}
The field of quantitative trading is extremely high-stakes. The hallucinations and incomplete inputs of LLMs cannot be used to trade in an environment where there are millions of dollars on the line. In order for these models to be used practically in this field, they must be able to understand and apply fundamental trading concepts and basic market dynamics, such as:
\begin{itemize}[leftmargin=*]
    \item Trading rules and order execution
    \item Implementing strategies in backtests
    \item Understanding market and risk data
\end{itemize}
\paragraph{}
The existing evaluations of large language models in finance focus on high-level tasks such as summarizing earnings calls, evaluating company fundamentals, modeling cash flows, or predicting sentiment of headlines. There has been little research that evaluates models on their ability to assist traders on a day-to-day basis.
\paragraph{}
This paper takes a step toward closing that gap by introducing \marketbench, a benchmark that evaluates:
\begin{quote}
\emph{Given a description of a trading strategy and market data, can a large language model construct a backtest whose output metrics match those of a verifiable implementation?}
\end{quote}
\paragraph{}
We design \marketbench{} around three strategies that capture some of the fundamental aspects of market dynamics:
\begin{enumerate}[leftmargin=*]
    \item \textbf{Scheduled trading} on a single stock (NASDAQ: MSFT), focusing on order-book interaction, position tracking, and P\&L accounting.
    \item \textbf{Pairs trading} on (NASDAQ: KO) and (NASDAQ: PEP), stressing spread computation, z-score-based entry/exit rules, and joint capital management across multiple symbols.
    \item \textbf{Options delta hedging} on (NASDAQ: MSFT), focusing on hedging deltas from an external options portfolio using the underlying stock.
\end{enumerate}

\section{Related Work}

\paragraph{Financial large language models.}
There has been increased work to build domain-specific large language models for finance. BloombergGPT, for example, is an early proprietary model trained on a large mixture of general and financial data to support tasks such as sentiment analysis, news classification, and question answering within the Bloomberg ecosystem \citep{wu2023bloomberggpt}. On the other hand, FinGPT proposes an open-source pipeline for financial large language models that emphasizes automatic data pulling and lightweight fine-tuning so that models can be continually adapted to new market information \citep{yang2023fingpt}. Furthermore, the PIXIU framework introduces an evaluation suite that covers multiple financial NLP and prediction tasks, providing one of the first publicly available financial LLM + benchmarks \citep{xie2023pixiu}. More recently, FinBen and Open-FinLLMs continue in this direction by covering a broad range of financial tasks and multimodal benchmarks and models (text, tables, time series, and charts), respectively \citep{xie2024finben,xie2024openfinllms}. In addition, FinanceQA also demonstrates that models fail 60\% of realistic tasks at hedge funds and other financial institutions \cite{mateega2025financeqabenchmarkevaluatingfinancial}. Most of the current work in this area evaluates high-level language tasks such as classification, extraction, and textual analysis.

\paragraph{Financial benchmarks for LLMs.}
There have been several benchmarks on evaluating large language models in financial domains. FinEval targets Chinese financial domain knowledge through thousands of multiple-choice questions which span academic finance along with industry practice. In addition, CFinBench builds a comprehensive Chinese financial benchmark which tests professional qualification exams and roles like tax consultants and securities analysts \citep{nie2024cfinbench}. FinEval-KR further separates knowledge versus reasoning ability and introduces separate metrics and datasets to study both at the same time \citep{dou2025finevalkr}. BizFinBench evaluates practical, business driven applications like calculation, reasoning, and information extraction \citep{lu2025bizfinbench}. While these benchmarks provide broad coverage of financial reasoning and understanding, they do not test how these models could be practically used in a quantitative trader's day to day life.

\paragraph{Code-generation and program-synthesis benchmarks.}
On the code side, \marketbench{} has similar aspects to general purpose code generation benchmarks. HumanEval evaluates models trained on code by asking them to create function bodies that pass unit tests \citep{chen2021codex}. MBPP (Mostly Basic Programming Problems) measures creation on short, natural language problems with unit test evaluation as well \citep{austin2021programsynthesis}. DS-1000 targets data science code generation in realistic settings by using problems from StackOverflow spanning seven Python libraries. It highlights the inconsistency of model generated code for data science tasks \citep{lai2022ds1000}. Furthermore, SWE-Bench tested models on 2,294 different software engineering problems \cite{jimenez2024swebenchlanguagemodelsresolve}. Current state-of-the-art models can solve $\sim$70\% of the latest SWE-bench problems. These benchmarks stress code correctness but are not applied to financial market structure. \marketbench{} instead focuses on domain-specific backtesting and market mechanics, evaluating models both on code reliability and on error of P\&L, positions, and risk metrics when compared to a verifiable implementation.

\section{Benchmark Design}

\subsection{Data Collection}
All of the datasets were either generated synthetically through a random process or obtained from \citep{databento_us_equities}. We preprocessed the datasets by randomizing the volume available at each price level and only using the top 3 levels available. This was done to ensure that the models tracked the liquidity that trades remove from the book and whether they persisted that liquidity correctly. Furthermore, the options delta dataset for Strategy 3 was generated using a simple random walk.

\subsection{High-Level Strategy Descriptions}

\paragraph{Strategy 1: Scheduled market-order execution on MSFT.}
Strategy~1 uses data from Databento's Market-by-price L10 data for Microsoft (MSFT). At pre-specified timestamps in the data, the strategy sends a market order to either buy or sell various quantities of MSFT. Each market order takes liquidity from the current book, net of previous trades at that price level and potentially from several price levels at once. 
\paragraph{}
The strategy tracks:
\begin{itemize}[leftmargin=*]
    \item Cash and MSFT position,
    \item Realized P\&L using FIFO accounting,
    \item Unrealized P\&L based on raw-book mid-prices,
    \item An equity curve and maximum drawdown,
    \item Synthetic-book statistics such as total size available and bid/ask VWAP post model trades.
\end{itemize}

\paragraph{Strategy 2: Pairs mean-reversion on Coke and Pepsi stock.}
Strategy~2 is a pairs trading strategy between KO and PEP which uses L10 order book data from both. At each new book update for either symbol, the strategy calculates mid-prices for both symbols and creates a spread between them as a linear combination of the two. A rolling history of the spread is then used to calculate a mean and the z-score of the current value.
\paragraph{}
The strategy has a position state (flat, long-spread, or short-spread). When the z-score exceeds an entry threshold, the strategy enters a mean-reversion position by buying one leg and selling the other. Positions are then flattened when the z-score reverts toward zero below an exit threshold. Additional features include:
\begin{itemize}[leftmargin=*]
    \item A cooldown mechanism that prevents quick re-entry in the same direction,
    \item A shared capital account for both symbols,
    \item Synthetic books and VWAP tracking per symbol,
    \item Immediate-or-cancel limit orders priced from synthetic mid and spread.
\end{itemize}

\paragraph{Strategy 3: Options delta hedging on MSFT.}
Strategy~3 utilizes MSFT order book data alongside a predefined option delta time series from a ``separate'' strategy. At regular time intervals, this strategy evaluates the current net delta and trades a portion of that delta to get flat. A minimum time difference between hedges is also enforced.
\paragraph{}
Hedge trades use fill-or-kill limit orders, where the limit price is set from synthetic mid-price and book spread. Each order experiences a fixed exchange delay before execution. As in the other strategies, a synthetic book persists consumed liquidity, and we track:
\begin{itemize}[leftmargin=*]
    \item Stock position and options delta,
    \item Net delta of the combined portfolio,
    \item Realized and unrealized P\&L from stock trades,
    \item Equity and maximum drawdown.
\end{itemize}
\paragraph{}
All three strategies required the models to track and reserve the liquidity they removed from the book through simulated trades. This is done by reserving these prices and creating a ``synthetic book'' which nets the raw order book data and the consumed liquidity. Furthermore, Strategies~2 and~3 also include a delay between submitting an order and hearing back from the exchange, mirroring the real world.

\subsection{Prompt Design}
The prompt for each strategy included relevant information describing the strategy along with information for the input and output datasets. The column names of the input dataset were explicitly detailed. Reasoning and thinking were enabled for the models that support it, and the temperature was set to 0.0. A time limit of 10 minutes was enforced for the model to give a valid response. Furthermore, the allowed packages were: \texttt{pandas}, \texttt{numpy}, \texttt{pathlib}, \texttt{datetime}, \texttt{collections}, \texttt{typing}, \texttt{statistics}, \texttt{math}, \texttt{sys}, \texttt{os}, and \texttt{dataclasses}.

\subsection{Evaluation Structure}

For each strategy $s \in \{1,2,3\}$, we define five distinct rounds $r \in \{1,\dots,5\}$. The results from each round are averaged to create a final result for that model. Each round corresponds to a specific input dataset, parameter configuration, and reference implementation. Every model $m$ is evaluated on all $3 \times 5 = 15$ (strategy, round) combinations.
\paragraph{}
For each $(s,m,r)$, we sample up to $K=3$ independent, one-shot attempts. An attempt refers to the model receiving the input data alongside the prompt and outputting code. This code is then checked and run to ensure successful execution. If execution succeeds, we record:
\begin{itemize}[leftmargin=*]
    \item \texttt{status} = \texttt{SUCCESS},
    \item The resulting average MAE between model-generated and reference metrics.
\end{itemize}
If execution fails (e.g., due to syntax errors, missing fields, or runtime assertions), we record:
\begin{itemize}[leftmargin=*]
    \item \texttt{status} = \texttt{FAILED},
    \item The error trace
\end{itemize}

\section{Evaluation Metrics and Protocol}

\subsection{Per-Attempt Metrics}

For each attempt, we compute \texttt{average\_mae} defined as the mean absolute error between the vector of reference metrics $y$ and the vector of model-generated metrics $\hat{y}$:
\begin{equation}
    \text{MAE}(y, \hat{y}) = \frac{1}{d} \sum_{i=1}^{d} \left| y_i - \hat{y}_i \right|,
\end{equation}
where $d$ is the number of scalar metrics (e.g., total P\&L, max drawdown, etc.) produced by the backtester for the final state of the simulation.
\paragraph{}
For attempts that do not produce comparable outputs (failed attempts), we assign a strategy specific \textbf{baseline MAE} by repeating the \emph{initial} (first row) metrics for all timestamps and computing its MAE against the true metrics time series. This baseline corresponds to not building a backtest at all as it repeats what was initially given.

\subsection{Per-Round Aggregation}

For each $(s,m,r)$ (strategy, model, round) triple, we aggregate over attempts as follows:
\begin{align}
    \text{best\_mae}_{s,m,r} &= \min \{ \text{MAE} : \text{status} = \texttt{SUCCESS} \},
\end{align}
and if a round is never solved, we set \texttt{best\_mae} to the strategy-specific baseline MAE and treat the round as unsolved for executability counting.

\subsection{Per-Strategy Metrics}

For each model $m$ and strategy $s$, we define:
\begin{align}
    \text{Mean MAE}_{s,m} &= \frac{1}{5} \sum_{r=1}^{5} \text{best\_mae}_{s,m,r}, \\
    \text{Best Run MAE}_{s,m} &= \min_{r \in \{1,\dots,5\}} \text{best\_mae}_{s,m,r}, \\
    \text{Executable Passes}_{s,m} &= \sum_{r=1}^{5} \mathbf{1}\{\text{at least one attempt is \texttt{SUCCESS}}\}.
\end{align}
Here, \text{Mean MAE} averages numerical error across all rounds, including baseline-filled failures; \text{Best Run MAE} captures the best numerical fidelity achieved in any round; and \text{Executable Passes} counts how many rounds produced an executable backtest.

\subsection{Overall Metrics and Ranking}

To obtain an overall view across strategies, we aggregate per model over all $15$ (strategy, round) combinations:
\begin{align}
    \text{Mean MAE}_{m} &= \frac{1}{15} \sum_{s=1}^{3} \sum_{r=1}^{5} \text{best\_mae}_{s,m,r}, \\
    \text{Best Run MAE}_{m} &= \min_{s \in \{1,2,3\},\, r \in \{1,\dots,5\}} \text{best\_mae}_{s,m,r}, \\
    \text{Executable Passes}_{m} &= \sum_{s=1}^{3} \sum_{r=1}^{5} \mathbf{1}\{\text{at least one attempt is \texttt{SUCCESS}}\}.
\end{align}
\paragraph{}
Within each strategy (and overall), we rank models by: Mean MAE.
\textbf{All scalar figures reported in the tables and text below (MAE and executable passes) are rounded to two decimal places.}

\section{Experimental Results}

\subsection{Strategy-Level Results}

Tables~\ref{tab:strategy1-results}--\ref{tab:strategy3-results} show the per-strategy rankings for all models, sorted from best to worst. Each table lists mean MAE, best run MAE, and executable passes.

\subsubsection{Strategy 1: Single-Stock Scheduled Execution}

\begin{table}[H]
\centering
\begin{tabular}{lccc}
\toprule
Model & Mean MAE & Best Run MAE & Executable Passes \\
\midrule
gemini-3-pro-preview         & 14.83   & 14.83   & 5 \\
claude-sonnet-4.5            & 16.36   & 16.35   & 5 \\
claude-opus-4.5              & 111.40  & 7.18    & 3 \\
grok-4                       & 121.85  & 7.22    & 4 \\
deepseek-v3.2                & 139.19  & 7.26    & 5 \\
command-a                    & 184.90  & 153.10  & 1 \\
nova-premier-v1              & 242.34  & 153.10  & 2 \\
mistral-large-2512           & 361.87  & 23.01   & 5 \\
gpt-5.1-codex-max            & 844.74  & 0.002   & 5 \\
gpt-5.2                      & 1{,}545.89 & 153.10  & 5 \\
qwen3-max                    & 2{,}015.66 & 16.39   & 5 \\
llama-3.1-nemotron-ultra     & 2{,}511.13 & 111.63  & 3 \\
llama-4-maverick             & 4{,}137.62 & 170.06  & 5 \\
\bottomrule
\end{tabular}
\caption{Strategy 1: Scheduled execution on MSFT. Eight of thirteen models solve all five rounds (Executable Passes = 5).}
\label{tab:strategy1-results}
\end{table}
\paragraph{}
Strategy~1 is the easiest task. No model fails all five rounds, and the average Executable Passes across all models is $4.08$ out of $5$. However, the errors of the outputs vary considerably.

\begin{itemize}[leftmargin=*]
    \item GPT-5.1 Codex-Max achieves the lowest best-run MAE (0.002) with Executable Passes = 5, though its mean MAE (844.74) is higher due to variance across rounds.
    \item Gemini 3 Pro and Claude Sonnet combine reliability (Executable Passes = 5) with very small mean MAE (14.83 and 16.36, respectively).
    \item DeepSeek V3.2 and Mistral-Large-2512 also solve all rounds but with higher MAE (139.19 and 361.87), while Llama-4 Maverick has substantially larger error (4{,}137.62) despite Executable Passes = 5.
    \item Qwen3 Max, GPT-5.2, Amazon Nova Premier, Nvidia Nemotron, and Cohere Command-A frequently produce executable backtests but with errors in the hundreds to thousands.
\end{itemize}
\paragraph{}
Strategy~1's results show that even relatively simple single-asset execution logic for market orders requires more reasoning than models can currently execute. Small mistakes in lot tracking or synthetic-book handling accumulate into large discrepancies.

\subsubsection{Strategy 2: Pairs Mean-Reversion on COKE/PEP}

\begin{table}[H]
\centering
\begin{tabular}{lccc}
\toprule
Model & Mean MAE & Best Run MAE & Executable Passes \\
\midrule
gemini-3-pro-preview         & 52.22   & 52.22   & 5 \\
gpt-5.2                      & 107.02  & 48.10   & 5 \\
deepseek-v3.2                & 132.77  & 125.87  & 3 \\
nova-premier-v1              & 133.10  & 133.10  & 0 \\
claude-opus-4.5              & 133.86  & 85.56   & 2 \\
gpt-5.1-codex-max            & 136.97  & 89.43   & 5 \\
claude-sonnet-4.5            & 193.32  & 70.82   & 5 \\
mistral-large-2512           & 228.90  & 133.10  & 5 \\
grok-4                       & 309.37  & 119.25  & 4 \\
llama-4-maverick             & 605.26  & 131.92  & 2 \\
command-a                    & 1{,}093.36 & 133.10  & 2 \\
llama-3.1-nemotron-ultra     & 2{,}268.11 & 133.10  & 2 \\
qwen3-max                    & 408{,}991{,}460.86 & 100.14  & 5 \\
\bottomrule
\end{tabular}
\caption{Strategy 2: Pairs mean-reversion on COKE and PEPSI.}
\label{tab:strategy2-results}
\end{table}
\paragraph{}
Strategy~2 is more structurally challenging. Amazon Nova Premier never produces a successful attempt (Executable Passes = 0), and several other models solve only two of the five rounds. On average, models achieve Executable Passes of 3.46. The errors of the model outputs also show a wide spread:
\begin{itemize}[leftmargin=*]
    \item Gemini 3 Pro again stands out, with Executable Passes = 5 and the lowest mean MAE of 52.22.
    \item GPT-5.2 and GPT-5.1 Codex-Max achieve low MAE (107.02 and 136.97, respectively), and both have Executable Passes = 5.
    \item Claude Sonnet and Mistral-Large-2512 both solve all rounds with moderate MAE (193.32 and 228.90).
    \item Qwen3 Max attains Executable Passes = 5 but has an MAE on the order of $4.09 \times 10^8$, reflecting extreme divergence in logic from the verifiable backtester.
\end{itemize}
\paragraph{}
This strategy stresses multi-asset state management along with strict entry and exit rules. Errors in z-score computation, sizing of legs, or shared capital can cause large discrepancies from the intended behavior.

\subsubsection{Strategy 3: Delta Hedging with MSFT}

\begin{table}[H]
\centering
\begin{tabular}{lccc}
\toprule
Model & Mean MAE & Best Run MAE & Executable Passes \\
\midrule
gpt-5.2                      & 1{,}369.59   & 1{,}365.42  & 5 \\
grok-4                       & 1{,}482.33   & 1{,}013.33  & 5 \\
gemini-3-pro-preview         & 4{,}595.52   & 1{,}013.99  & 5 \\
gpt-5.1-codex-max            & 10{,}496.40  & 1{,}370.72  & 5 \\
deepseek-v3.2                & 14{,}724.36  & 1{,}127.14  & 5 \\
claude-sonnet-4.5            & 16{,}157.31  & 805.16      & 5 \\
claude-opus-4.5              & 18{,}945.34  & 1{,}370.72  & 2 \\
llama-4-maverick             & 21{,}279.52  & 20{,}418.58 & 1 \\
nova-premier-v1              & 21{,}345.75  & 21{,}345.75 & 0 \\
llama-3.1-nemotron-ultra     & 21{,}345.75  & 21{,}345.75 & 0 \\
command-a                    & 21{,}369.58  & 19{,}852.90 & 2 \\
mistral-large-2512           & 63{,}683.24  & 1{,}351.55  & 4 \\
qwen3-max                    & 329{,}700.43 & 1{,}087.64  & 5 \\
\bottomrule
\end{tabular}
\caption{Strategy 3: Delta-hedging MSFT against options delta. MAE spans several orders of magnitude.}
\label{tab:strategy3-results}
\end{table}

Strategy~3 is the most complex and numerically the harshest task. Two models (Amazon Nova Premier and Nvidia Nemotron) never solve a round (Executable Passes = 0), while seven models achieve Executable Passes = 5. Average Executable Passes across models is 3.38, and mean MAE spans several orders of magnitude.
\begin{itemize}[leftmargin=*]
    \item GPT-5.2 achieves the lowest mean MAE (1{,}369.59) with Executable Passes = 5.
    \item Grok 4 also performs well, with Executable Passes = 5 and a mean MAE of 1{,}482.33.
    \item GPT-5.1 Codex-Max, Claude Sonnet, and DeepSeek V3.2 solve all rounds but with MAE in the range of 10{,}496.40--16{,}157.31.
    \item Qwen3 Max again shows extreme numerical divergence (329{,}700.43 MAE) despite consistent output.
\end{itemize}
The sensitivity of this strategy comes from the interaction between the options delta series, the timing and size of hedge orders, and the fill-or-kill order execution type along with exchange delay. Small conceptual mistakes in how deltas are aggregated or how hedges are throttled can accumulate into very large mismatches in P\&L and net delta over the course of the simulation.

\subsection{Strategy Difficulty}

We can summarize strategy difficulty by averaging metrics across models (excluding NaN MAE values):
\begin{itemize}[leftmargin=*]
    \item \textbf{Strategy 1}: Average Executable Passes $= 4.08$ and average mean MAE $= 9.42 \times 10^{2}$. Excluding Qwen3 Max, the average mean MAE drops to $8.53 \times 10^{2}$.
    \item \textbf{Strategy 2}: Average Executable Passes $= 3.46$ and average mean MAE $= 3.15 \times 10^{7}$, dominated by the extreme outlier from Qwen3 Max. Excluding Qwen3 Max, the average mean MAE falls to $4.50 \times 10^{2}$.
    \item \textbf{Strategy 3}: Average Executable Passes $= 3.38$ and average mean MAE $= 4.20 \times 10^{4}$. Excluding Qwen3 Max, the average mean MAE drops to $1.81 \times 10^{4}$.
\end{itemize}
\paragraph{}
These results are unsurprising; Strategy~1 is the easiest to solve numerically and consistently, with each subsequent strategy becoming harder to solve accurately and reliably.

\subsection{Overall Model Comparison}

Table~\ref{tab:overall-results} presents overall metrics averaged across all three strategies, again sorted from best to worst by mean MAE.

\begin{table}[H]
\centering
\begin{tabular}{lccc}
\toprule
Model & Mean MAE & Best Run MAE & Executable Passes \\
\midrule
grok-4                       & 443.24     & 7.22    & 13 \\
gpt-5.2                      & 969.39     & 48.10   & 15 \\
gemini-3-pro-preview         & 1{,}744.27 & 14.83   & 15 \\
gpt-5.1-codex-max            & 4{,}242.93 & 0.002   & 15 \\
deepseek-v3.2                & 4{,}575.64 & 7.26    & 13 \\
claude-sonnet-4.5            & 5{,}126.76 & 16.35   & 15 \\
claude-opus-4.5              & 6{,}039.62 & 7.18    & 7 \\
command-a                    & 6{,}562.11 & 133.10  & 5 \\
nova-premier-v1              & 7{,}740.26 & 133.10  & 2 \\
llama-3.1-nemotron-ultra     & 9{,}674.21 & 111.63  & 5 \\
llama-4-maverick             & 10{,}202.18 & 131.92 & 8 \\
mistral-large-2512           & 30{,}605.53 & 23.01  & 14 \\
qwen3-max                    & 159{,}144{,}490.12 & 16.39 & 15 \\
\bottomrule
\end{tabular}
\caption{Overall \marketbench{} performance across all three strategies.}
\label{tab:overall-results}
\end{table}

These overall results highlight several themes:
\begin{itemize}[leftmargin=*]
    \item \textbf{Reliability vs.\ accuracy.} Qwen3 Max achieves perfect overall Executable Passes ($15$), but its overall mean MAE is enormous ($\approx 1.59 \times 10^{8}$) due to extreme divergence on Strategy~2. In contrast, Grok 4 achieves the lowest overall mean MAE (443.24) despite not solving all rounds, and GPT-5.2 combines perfect executability with low mean MAE of 969.39.
    \item \textbf{Inconsistency.} Cohere Command-A, Amazon Nova Premier, and Nvidia Nemotron frequently fail to produce executable backtests on the harder strategies, leading to low overall Executable Passes even when their MAE on the few solved rounds is not uniformly bad. Furthermore, almost every model showed large variations within the rounds themselves.
\end{itemize}

\section{Discussion and Failure Modes}

\subsection{Structural vs.\ Semantic Failures}

The logs reveal two different types of failures:

\paragraph{Structural failures.}
These occur when the model-generated output cannot be executed to produce valid, comparable results.
\begin{itemize}[leftmargin=*]
    \item Incorrect or missing function signatures,
    \item Invalid references to columns or fields in the market data,
    \item Inconsistent types in intermediate calculations.
\end{itemize}

\paragraph{Semantic failures.}
These occur when the backtest output by the model runs but does not faithfully implement the intended logic. The resulting MAE is large even though the backtester outputs metrics. For example:
\begin{itemize}[leftmargin=*]
    \item Miscomputing spreads or z-scores in Strategy~2 (for example, wrong hedge ratios between COKE and PEPSI or using wrong rolling windows),
    \item Processing the options delta stream or hedge timing in Strategy~3 incorrectly, or interpreting delta signs incorrectly,
    \item Ignoring capital constraints or misapplying FIFO lot accounting in Strategy~1.
\end{itemize}
Models that are consistent but logically unsound tend to show high Executable Passes with large mean MAE, as seen with Qwen3 Max, Mistral-Large-2512, and, to a smaller extent, DeepSeek V3.2.

\subsection{Implications for Real-World Use}

Even for introductory tasks, \marketbench{} shows that:
\begin{itemize}[leftmargin=*]
    \item High success rates in model output do not guarantee accuracy.
    \item Multi-asset interactions and hedging risk expose nontrivial weaknesses in the reasoning and understanding of large language models in the trading field.
    \item There is large variance in the outputs of the same model across different rounds, as the best-run MAE can differ substantially from the mean MAE for each strategy--model pair.
    \item Current models can only be used as a coding supplement instead of synthesizing trading ideas and strategies.
\end{itemize}
\paragraph{}
Currently, large language models struggle with understanding and implementing even basic quantitative trading strategies. Using model output as a drop-in replacement for the work of quants would be extremely risky. In the future, with better training and reasoning from trading data, there may be a path for models to become more useful.

\section{Limitations and Future Work}

\paragraph{Scope of strategies.}
\marketbench{} currently includes only three strategies, each focusing on a different aspect of market dynamics. The benchmark does not yet cover options pricing, multi-asset portfolios beyond pairs, intraday inventory risk limits, or transaction-cost-sensitive execution tactics.

\paragraph{Metric scaling.}
Our evaluation uses unnormalized MAE on absolute metrics, which can produce very large values for some strategies. While this reflects genuine numerical divergence, it complicates cross-strategy comparisons. Future work could incorporate relative errors, correlations of P\&L paths, or risk-adjusted performance metrics to provide additional information and evaluation.

\paragraph{Strategy explanation.}
In the future, we hope large language models can be applied in the trading industry to identify drawbacks in strategies and pinpoint why a strategy is losing or gaining on certain trades. However, this would require the model to understand and be able to apply the strategy itself.

\section{Conclusion}
We present \marketbench{}, a benchmark for evaluating large language models on introductory quantitative trading tasks that require both accuracy and consistency. By analyzing 329 total attempts across three unique trading strategies, we find that:
\begin{itemize}[leftmargin=*]
    \item Current models lack the capabilities to simulate and understand even basic trading strategies.
    \item Many models can reliably produce executable backtests on simpler strategies, but a subset fails catastrophically on more complex ones.
    \item Model error varies widely, especially on the most realistic Strategy~3, where small implementation differences can generate huge P\&L and risk discrepancies.
\end{itemize}
\paragraph{}
We hope \marketbench{} can serve as a foundation for future work on large language models that are not just capable of describing strategies but also implementing them in a way that demonstrates deep understanding of market dynamics, risk, and trading mechanics.

\bibliography{references}

@article{wu2023bloomberggpt,
  author  = {Wu, Shijie and others},
  title   = {BloombergGPT: A Large Language Model for Finance},
  journal = {arXiv preprint arXiv:2303.17564},
  year    = {2023},
  url     = {https://arxiv.org/abs/2303.17564}
}

@article{yang2023fingpt,
  author  = {Yang, Hongyang and Liu, Xiao-Yang and Wang, Christina Dan},
  title   = {{FinGPT}: Open-Source Financial Large Language Models},
  journal = {FinLLM Symposium at IJCAI 2023},
  year    = {2023},
  url     = {https://arxiv.org/abs/2306.06031}
}

@article{xie2023pixiu,
  author  = {Xie, Qianqian and others},
  title   = {{PIXIU}: A Large Language Model, Instruction Data and Evaluation Benchmark for Finance},
  journal = {arXiv preprint arXiv:2306.05443},
  year    = {2023},
  url     = {https://arxiv.org/abs/2306.05443}
}

@inproceedings{xie2024finben,
  author    = {Xie, Qianqian and others},
  title     = {{FinBen}: A Holistic Financial Benchmark for Large Language Models},
  booktitle = {Advances in Neural Information Processing Systems 37 (Datasets and Benchmarks Track)},
  year      = {2024},
  url       = {https://arxiv.org/abs/2402.12659}
}

@article{xie2024openfinllms,
  author  = {Xie, Qianqian and others},
  title   = {Open-{FinLLMs}: Open Multimodal Large Language Models for Financial Applications},
  journal = {arXiv preprint arXiv:2408.11878},
  year    = {2024},
  url     = {https://arxiv.org/abs/2408.11878}
}

@article{nie2024cfinbench,
  author  = {Nie, Ying and others},
  title   = {{CFinBench}: A Comprehensive Chinese Financial Benchmark for Large Language Models},
  journal = {arXiv preprint arXiv:2407.02301},
  year    = {2024},
  url     = {https://arxiv.org/abs/2407.02301}
}

@article{dou2025finevalkr,
  author  = {Dou, Shaoyu and others},
  title   = {{FinEval-KR}: A Financial Domain Evaluation Framework for Large Language Models' Knowledge and Reasoning},
  journal = {arXiv preprint arXiv:2506.21591},
  year    = {2025},
  url     = {https://arxiv.org/abs/2506.21591}
}

@article{lu2025bizfinbench,
  author  = {Lu, Guilong and others},
  title   = {{BizFinBench}: A Business-Driven Real-World Financial Benchmark for Evaluating {LLMs}},
  journal = {arXiv preprint arXiv:2505.19457},
  year    = {2025},
  url     = {https://arxiv.org/abs/2505.19457}
}

@misc{databento_us_equities,
  author       = {{Databento Inc.}},
  title        = {Databento {US} Equities},
  howpublished = {\url{https://databento.com/portal/catalog/us-equities}},
  year         = {2025}
}

@misc{jimenez2024swebenchlanguagemodelsresolve,
  title         = {{SWE-bench}: Can Language Models Resolve Real-World {GitHub} Issues?},
  author        = {Jimenez, Carlos E. and Yang, John and Wettig, Alexander and Yao, Shunyu and Pei, Kexin and Press, Ofir and Narasimhan, Karthik},
  year          = {2024},
  eprint        = {2310.06770},
  archivePrefix = {arXiv},
  primaryClass  = {cs.CL},
  url           = {https://arxiv.org/abs/2310.06770}
}

@misc{mateega2025financeqabenchmarkevaluatingfinancial,
  title         = {{FinanceQA}: A Benchmark for Evaluating Financial Analysis Capabilities of Large Language Models},
  author        = {Mateega, Spencer and Georgescu, Carlos and Tang, Danny},
  year          = {2025},
  eprint        = {2501.18062},
  archivePrefix = {arXiv},
  primaryClass  = {cs.LG},
  url           = {https://arxiv.org/abs/2501.18062}
}

@article{chen2021codex,
  author  = {Chen, Mark and others},
  title   = {Evaluating Large Language Models Trained on Code},
  journal = {arXiv preprint arXiv:2107.03374},
  year    = {2021},
  url     = {https://arxiv.org/abs/2107.03374}
}

@article{austin2021programsynthesis,
  author  = {Austin, Jacob and others},
  title   = {Program Synthesis with Large Language Models},
  journal = {arXiv preprint arXiv:2108.07732},
  year    = {2021},
  url     = {https://arxiv.org/abs/2108.07732}
}

@article{lai2022ds1000,
  author  = {Lai, Zheng and others},
  title   = {{DS-1000}: A Natural and Reliable Benchmark for Data Science Code Generation},
  journal = {arXiv preprint arXiv:2211.11501},
  year    = {2022},
  url     = {https://arxiv.org/abs/2211.11501}
}

\end{document}